\documentclass{article}
\usepackage[utf8]{inputenc}
\usepackage[arabic,english]{babel}
\DeclareMathSymbol{\mtimes}{\mathbin}{symbols}{"02}
\usepackage{amsmath,amssymb,scalefnt}
\usepackage{makecell}
\usepackage{tikz}
\usepackage{color,soul}
\usepackage{hyperref}
\usepackage{enumerate}
\usepackage{pgfplots}
\usepackage{apacite}
\usepackage[most]{tcolorbox}
\usepackage{caption}
\usepackage{float}
\usepackage{subcaption}
\usepackage[left,modulo]{lineno}
\usetikzlibrary{matrix,calc}
\usetikzlibrary{arrows,positioning,fit,shapes,calc,shadows,shapes.misc}
\pgfplotsset{width=10cm,compat=1.11}
\title{Hidden Markov Based Mathematical Model dedicated to Extract Ingredients from Recipe Text}
\usepackage[symbol]{footmisc}
\def\correspondingauthor{\footnote{Corresponding author: ziedbaklouti@outlook.com}}
\def\university{\footnote{MSc computer science and mathematics, Paris University}}
\author{Zied Baklouti\correspondingauthor{} \university{}}
\date{March 2021}
\tikzset{
every transaction/.style = {fill=white!100},
transaction1/.style = {starburst, draw , aspect=2, align=center, inner sep=1pt},
transaction4/.style = {starburst, draw , aspect=2, align=center, inner sep=1pt},
transaction2/.style = {double copy shadow={shadow xshift=2ex,shadow yshift=2ex},
transaction3/.style = {shape=rounded rectangle, shape example, inner xsep=1.5cm, inner ysep=1cm},
fill=white!20,draw=black,thick},
transaction/.style = {diamond, draw , aspect=2, align=center, inner sep=1pt},
every actor role/.style = {},
actor role/.style = {rectangle, draw=black!80, ultra thick,
    minimum size = 6mm, every actor role},
composite actor role/.style = {fill=white!50, actor role},
composite/.style = {fill=white!100, actor role},
elementary actor role/.style = {fill=white!100, actor role},
initiator/.style = {-},
executor/.style = {<-, >=squarea},
system/.style = {rectangle, fill=white!100, ultra thick, draw=black!80,
            minimum height=60mm, minimum width=4cm,outer sep=0pt}}
\begin{document}

\maketitle
	\begin{abstract}
Natural Language Processing (NLP) is a branch of artificial intelligence that gives machines the ability to decode human languages. Part-of-speech tagging (POS tagging) is a pre-processing task that requires an annotated corpus. Rule-based and stochastic methods showed remarkable results for POS tag prediction. On this work, I performed a mathematical model based on Hidden Markov structures and I obtained a high-level accuracy of ingredients extracted from text recipe with performances greater than what traditional methods could make without unknown words consideration. 
	\end{abstract}
\section{Introduction}

Artificial intelligence had shown a great progress in the recent years especially the deep learning branch where learning techniques have been improved very quickly. The combination of representation learning and deep learning have allowed the emerging of a new AI class called deep reinforcement learning.  

Deep Reinforcement learning tend to estimate value functions from experiments and simulations and using dynamic programming through Deep Reinforcement learning is an efficient way to build reactive strategies acting on instantaneous control. An algorithm which approves its performance by experience is an algorithm capable of avoiding his own mistakes through a combination of a strong memory fed by fresh helpful data and the ability to keep winning predictions after a long-term performance {\color{blue}\cite{barto1995learning}} {\color{blue}\cite{mnih2015human}}.

Neural Network can be considered as a dynamic Reinforcement Learning scheme where the layers are putted in a parallel way to have a cascaded transmission of the treated signal  {\color{blue}\cite{fukushima1982neocognitron}} {\color{blue}\cite{lecun1989backpropagation}} and where a prior knowledge is important to predict the output state of new observations.

Sequential modeling is a way to process data in natural language processing by maximizing awards after manipulating situation and producing resulting actions {\color{blue}\cite{vithayathil2020survey}} {\color{blue}\cite{lecun2015deep}}. A sequential model representation is influenced by its data representation and how tensors are trained to produce an optimal control {\color{blue}\cite{bengio2013representation}}

To improve the target learning task, transfer learning is used as a powerful technique to increase the value of the most probable cases inside a state matrix {\color{blue}\cite{boutsioukis2011transfer}}. Transferring the knowledge helps us to reduce the amount of data consumed and rely on feature engineering to reduce the noise caused by annotation errors and other tag-set anomalies in a context of multi-agent system.

Extracting ingredients automatically from a recipe text is an extremely useful activity especially when we want to analyze a massive data of text recipes. Rule-Based methods were implemented to extract information from unstructured recipe data {\color{blue}\cite{silva2019information}} Ingredients is not the only useful information we want to extract; in this work we are going to use Hidden Markov Models especially Viterbi algorithm with some modification to make it receiving two unique features: POS-tags and tokens, to predict ingredient states. 
	\section{Previous works}
    
Many previous works were interested in analyzing cuisine recipes, for example Sina Sajadmanesh {\color{blue}\cite{sajadmanesh2017kissing}} presented an analysis of the ingredients diversity around the word using an ingredient-based classifier to differentiate between recipes around the word based on its geographical identity. Sina Sajadmanesh {\color{blue}\cite{sajadmanesh2017kissing}} studied the diversity of ingredients in dishes with introduction of global diversity (the ability to have diversified ingredients between recipes) and local diversity (the ability to have diversified ingredients within a recipe). 

Other related work for culinary habits is Yong-Yeol Ahn paper {\color{blue}\cite{ahn2011flavor}} who introduced the notion of Flavor Network and tried to verify the Food Paring hypothesis introduced on the 90's by Heston Blumenthal and Francois Benzi. Flavor network as described by Ahn is a graph where the nodes are the ingredients extracted from recipes and weights are shared flavors between nodes. Food paring hypothesis is an indicator calculated after forming the Flavor Network to show if in a country or in a geographical part of the word we have tasty recipes or the ingredients do not have similar molecules. Tiago Simas {\color{blue}\cite{simas2017food}} introduced the notion of food bridging formed with semi-metric distances.

A group of scientists in a recent publication {\color{blue}\cite{van2021using}} developed a state of the art of the use of artificial intelligence and natural language processing in analyzing food recipes. In this article we can found collected references talking about the challenging part in collecting food and recipe data. For example, Ahnert {\color{blue}\cite{ahnert2013network}} presented the emergence of computational gastronomy in food science and its effect on culinary practices. Aiello and al {\color{blue}\cite{aiello2019large}} discovered what are the most important predictors in food responsible of three diseases in a population situated in London. {\color{blue}\cite{amato2020safeeat}} extracted ingredients from food text to alert readers from allergens presence in a recipe. I agree with {\color{blue}\cite{van2021using}} concerning how challenging to use IA in food domain and how it will resolve issues concerning the creation of a data driven analysis of nutrition. In our paper data extracted can be used in a phone application or a recommended system for people who want to take care of their health.

All previously cited researches on cuisine recipes need information extraction from text recipe to use it on graphical visualization and statistical analysis. Information extraction can be used manually by extracting ingredients indicated on recipes or automatically. The problem in automatic extraction is that information should be precise to have also precise analysis, for example some ingredients take only one word and others can take two or three words. Another problem on automatic information extraction is that some ingredients that take one word have in common some words with other ingredients that have more than one word which make automatic information extraction more difficult. I tried to develop a mathematical model dedicated to extract ingredients from text recipe written in Arabic language with precision higher than what traditional methods could make. According to Cutting {\color{blue}\cite{cutting1992practical}}, a Tagger must be robust that should deal with unknown words, efficient that can deal with large corpora, accurate that can tag with high accuracy, tunable that can deal with different corpora and reusable that take small efforts to re-target a new corpus. There are three types of POS Tagger: Taggers based on stochastic models, Taggers based on rules and Taggers Based on neural networks. On this work we will use Taggers based on HMM models. The use of POS tags as external features to solve NER problems was experimented by Zhou {\color{blue}\cite{zhou2002named}} but it was discarded because it showed bad results but our methodology and experiments demonstrate that using POS tags as external features is not a bad idea. This could be explained by the difference between our tokens and Zhou's tokens: tokens as defined by Zhou is a pair of word-feature and in our model, token is only a word from our corpus. 

\section{Training the dataset}

Hidden Markov Models are used in previous works to resolve named entity recognition (NER) problems. In our case we have a NER problem with one Named Entity to extract (Ingredients) and with various boundaries (one word, two words, three words). 

Our sentences belong to only one category of phrases on Arabic language: The noun phrase "\AR{الجمل الاسمية}". It is a kind of phrase that don't begin with verbs (phrase that don't describe an action). 

Noun phrases are constituted with 2 parts: The beginning of a sentence "\AR{المبتدا}" contains the object of the information, and the end of a sentence "\AR{الخبر}" containing the core information. 

{\color{blue}Table\ref{fig:tab1}} shows how the dataset is trained: 

\begin{table}[htbp] 

\begin{center} 

\begin{tabular}{|c|c|c|c|} 

\hline  

English translation & Tokens in arabic & Label & detected ingredients\\  

\hline  

a little of & \AR{رشة} & C & 0\\  

\hline  

salt & \AR{ملح} & D & 1\\  

\hline  

and & \AR{و} & J  & 0\\  

\hline  

black (\AR{اسود}) & \AR{فلفل} & E  & 1\\  

\hline  

pepper (\AR{فلفل}) & \AR{اسود} & F & 2\\  

\hline  

. & . & . & 0\\  

\hline

\end{tabular} 

		\centering 

\caption{Example of trained sentence in our trained corpus} 

\label{fig:tab1} 

\end{center} 

\end{table} 

When an ingredient is found on a corpus, we attribute 1 to the concerned token. If the information we want to extract takes more than one word we attribute 2 to the extra word completing the information. ${0,1,2,3}$ is the new list of tags used on the HMM tagger in the second phase of our ingredient extractor system. Our word dictionary is constituted with 807 words. The number of trained sentences is 1973. The number of POS tags appearing on the first layer of our ingredient extractor is 14 different POS tags. The sentences are describing ingredients constituting a cuisine recipe. {\color{blue}Table \ref{fig:tab1}} shows an example of a trained sentence. The entire trained dataset and the code used for this paper is available in this \href{https://github.com/Zied130/Hidden-Markov-Based-Mathematical-Model}{{\color{blue}gethub link}}\footnote{https://github.com/Zied130/Hidden-Markov-Based-Mathematical-Model}. 

\section{Methodology}

Our Ingredient extractor is constituted with two parts. In the first part we are going to predict part of speech tags of our sequences. On the second part we are going to use predicted Part of Speech as a grammatical feature to predict ingredients state. 

\subsection{Grammatical Feature Collection}

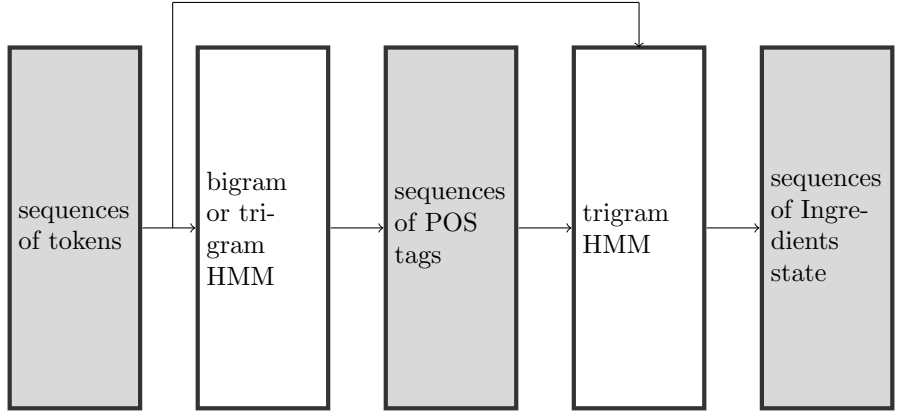
\begin{figure}[htbp]
\center
\begin{tikzpicture}[node distance=1cm, on grid]
    \node [composite actor role] (CA01) [ fill=gray!30,minimum height=48mm,text width=1.5cm] at ( -4,3) {sequences of tokens};
    \node [composite] (CA01) [minimum height=48mm,text width=1.5cm] at ( -1.5,3) {bigram or trigram HMM};
    \node [composite actor role] (CA01) [fill=gray!30,minimum height=48mm,text width=1.5cm] at ( 1,3) {sequences of POS tags};
    \node [composite] (CA01) [minimum height=48mm,text width=1.5cm] at ( 3.5,3) {trigram HMM};
    \node [composite actor role] (CA01) [fill=gray!30,minimum height=48mm,text width=1.5cm] at ( 6,3) {sequences of Ingredients state};
    \draw[->] ( -3.1,3) to ( -2.4,3);
    \draw[->] ( -0.6,3) to ( 0.1,3);
    \draw[->] (1.9,3) to ( 2.6,3);
    \draw[->] (4.4,3) to (5.1,3);
    \draw[-] ( -2.7,3) to ( -2.7,6);
    \draw[-] ( -2.7,6) to ( 3.5,6);
    \draw[->] ( 3.5,6) to ( 3.5,5.4);
\end{tikzpicture}
\caption{Methodology Diagram of our Ingredient Extractor}
\label{fig:fig1}
\end{figure}

In our Model we have observation sequences which are sentences containing the information we want to extract (Ingredients). The first part of our Ingredient Extractor will try to collect grammatical features by resolving the POS tagging problem. At the end of this part, we want to obtain the grammatical role for each token on each sequence. We experimented both first order HMM and second order HMM to predict ingredients state. We obtained strangely better results with first order HMM. This could be explained by the fact that in Arabic language a POS tag is organized in bi-gram manner. For example, if we have an E tag \AR{منعوت} we are certain to have an F tag \AR{نعت} because in Arabic a word tagged as an E tag \AR{منعوت} could be tagged as a noun but if the intention of the writer is to describe that noun with an adjective coming after it the exact tag to be used is an E tag \AR{منعوت}. 

\begin{table}[h!]
	 \begin{center}
	 \begin{tabular}{|c|c|c|}
\hline 
POS Tag (arabic) & POS Tag (english) & Label \\ 
\hline 
\AR{إسم} & Noun & A \\ 
\hline 
\AR{رقم} & Number & B \\ 
\hline 
\AR{إسم معرف} & Known Noun & C \\ 
\hline 
\AR{إضافة} & Noun after Known Noun & D \\ 
\hline 
\AR{منعوت} & Noun before the Adjective & E \\ 
\hline 
\AR{نعت} & Adjective & F \\ 
\hline 
\AR{حرف جر} & Preposition & G \\ 
\hline 
\AR{إسم مجرور} & Noun after preposition & H \\ 
\hline 
\AR{وحدة قيس} & unit of measure  & I \\ 
\hline 
\AR{واو العطف} & AND & J \\ 
\hline 
\AR{فعل مبني للمجهول} & Passive Verb & K \\ 
\hline 
\AR{المفعول المطلق} & Superlative & L \\ 
\hline 
\AR{أداةُ عَطْفٍ غير واو العطف} & OR & M \\ 
\hline 
. & . & . \\ 
\hline 
\end{tabular}
\caption{POS Tag-set used to annotate our corpus}
\label{tab:tab2}
\end{center}
\end{table}
The elements needed to define a first order HMM in order to perform a POS tag task are:
\begin{itemize}
\item N=14, the number of POS tags each token can take as shown in {\color{blue}Table \ref{tab:tab2}} we have 14 POS tag. We represent the sequence of POS tags by the sequence $T=\{t_{1},t_{2}... ,t_{14}\}$.
\item M=807, is the sequence of symbols generated by the stochastic procedure of the HMM model. In our case it is the list of tokens (words) constituting our lexicon.
\item Matrix A containing the state transition probability distribution. In our case it is the probability to move from one POS tag to another POS tag.
\begin{gather*}
	A=\begin{bmatrix}
	a_{11} & a_{12} & \cdots & a_{1n} \\
	a_{21} & a_{22} & \cdots & a_{2n} \\
	\vdots & \vdots & \ddots & \vdots \\	
	a_{n1} & a_{n2} & \cdots & a_{nn}
	\end{bmatrix}
	\end{gather*}
	\begin{gather*}
	a_{ij}=P(T_{r}=t_{j}/T_{r-1}=t_{i})
	\end{gather*}
\item Matrix B containing the observation symbol probability distribution. In our case it is the probability to emit a word $v_{k}$ given that the system is emitting one of the state of the sequence  $T=\{t_{1},t_{2}... ,t_{14}\}$.
\begin{gather*}
	B=\begin{bmatrix}
	b_{11} & b_{12} & \cdots & b_{1m} \\
	b_{21} & b_{22} & \cdots & b_{2m} \\
	\vdots & \vdots & \ddots & \vdots \\	
	b_{n1} & b_{n2} & \cdots & b_{nm}
	\end{bmatrix}
	\end{gather*}
\begin{gather*}
	b_{jk}=P(V_{r}=v_{k}/T_{r}=t_{j})
	\end{gather*}
\item $\pi$ a vector containing the probability distributions that a sentence begins with one of the elements of $T$
\begin{gather*}
	\pi=\begin{bmatrix}
	\pi_{1} & \pi_{2} & \pi_{3} & \hdots & \pi_{14} 
	\end{bmatrix}
	\end{gather*}
\end{itemize}

\subsection{Predicting Sequences of Ingredients State}

\subsubsection{Prediction Using First Order HMM}

Suppose that we want to use a first order HMM to predict ingredients state on the second part of our Ingredient Extractor without using grammatical features estimated on previous section, in this case the elements needed to define this HMM according to Rabiner{\color{blue}\cite{rabiner1989tutorial}}:
\begin{itemize}
\item N=4, the number of distinct states each token can take. In our case it is one element of the set $\Gamma={0,1,2,3}$
\item M=807, is the sequence of symbols generated by the stochastic procedure of the HMM model. In our case it is the list of tokens (words) constituting our lexicon.
\item Matrix A containing the state transition probability distribution. In our case it is the probability to move from one of the states on the set $\Gamma$ to another state on the same set.
\begin{gather*}
	A=\begin{bmatrix}
	a_{11} & a_{12} & \cdots & a_{1n} \\
	a_{21} & a_{22} & \cdots & a_{2n} \\
	\vdots & \vdots & \ddots & \vdots \\	
	a_{n1} & a_{n2} & \cdots & a_{nn}
	\end{bmatrix}
	\end{gather*}
	\begin{gather*}
	a_{ij}=P(\Gamma_{r}=\gamma_{j}/\Gamma_{r-1}=\gamma_{i})
	\end{gather*}
\item Matrix B containing the observation symbol probability distribution. In our case it is the probability to emit a word $v_{k}$ given that the system is emitting one of the states of the sequence $\Gamma$.
\begin{gather*}
	B=\begin{bmatrix}
	b_{11} & b_{12} & \cdots & b_{1m} \\
	b_{21} & b_{22} & \cdots & b_{2m} \\
	\vdots & \vdots & \ddots & \vdots \\	
	b_{n1} & b_{n2} & \cdots & b_{nm}
	\end{bmatrix}
	\end{gather*}
	\begin{gather*}
	b_{jv_k}=P(V_{r}=v_{k}/\Gamma_{r}=\gamma_{j})
	\end{gather*}
\item $\pi$ a vector containing the probability distributions that a sentence begins with one of the elements of $\Gamma$
\begin{gather*}
	\pi=\begin{bmatrix}
	\pi_{1} & \pi_{2} & \pi_{3} & \pi_{4} 
	\end{bmatrix}
	\end{gather*}
\end{itemize}
This model described above is a first-order hidden Markov model, we can call it also a bi-gram model to predict ingredients state.

\subsubsection{Prediction Using Full Second Order HMM}
Scott M.Thede{\color{blue}\cite{thede1999second}} used a full second-order Hidden Markov Model for his POS tagger. To constitute this trigram tagger from the previous bigram tagger, Scott M.Thede{\color{blue}\cite{thede1999second}} replaced bi-gram transition probability $a_{ij}$ with a trigram transition probability $a_{ijk}$ and replaced bi-gram lexical probability $b_{jv_k}$ with a trigram lexical probability $b_{ijv_k}$. If we want to use a full second-order hidden Markov model as described by Scott M.Thede{\color{blue}\cite{thede1999second}} the element needed to define this HMM are: 

\begin{itemize}
\item N=4, the number of distinct states each token can take.
\item M=807, number of tokens.
\item Matrix A shown on {\color{blue}figure \ref{fig:fig2}} containing the state transition probability distribution. For a trigram model this matrix is of dimensions  $N \mtimes N \mtimes N$.
\item Matrix B shown on {\color{blue}figure \ref{fig:fig3}} containing the observation symbol probability distribution. For trigram model this matrix is of dimensions  $N \mtimes N \mtimes M$
\item $\pi$ a vector containing the probability distributions that a sentence begins with one of the elements of $\Gamma$
\begin{gather*}
	\pi=\begin{bmatrix}
	\pi_{1} & \pi_{2} & \pi_{3} & \pi_{4} 
	\end{bmatrix}
	\end{gather*}
\end{itemize}

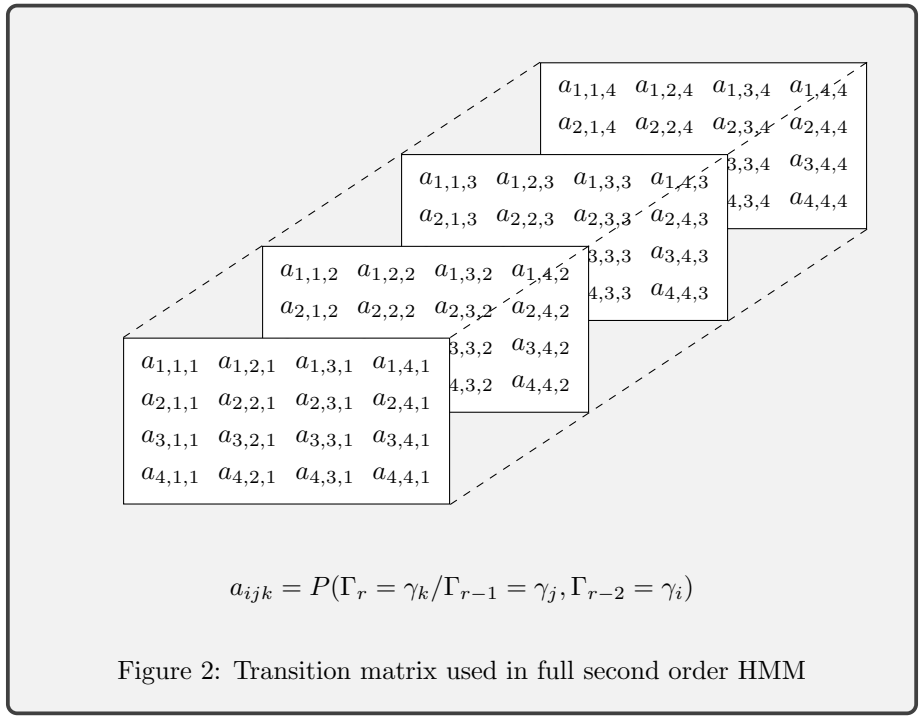
\begin{tcolorbox}
\begin{figure}[H]
\begin{tikzpicture}[every node/.style={anchor=north east,fill=white,minimum width=1mm,minimum height=2mm}]
\hspace*{1.0cm}
\matrix (mA) [draw,matrix of math nodes]
{
a_{1,1,4} & a_{1,2,4} & a_{1,3,4} & a_{1,4,4} \\
a_{2,1,4} & a_{2,2,4} & a_{2,3,4} & a_{2,4,4} \\
a_{3,1,4} & a_{3,2,4} & a_{3,3,4} & a_{3,4,4} \\
a_{4,1,4} & a_{4,2,4} & a_{4,3,4} & a_{4,4,4} \\
};

\matrix (mB) [draw,matrix of math nodes] at ($(mA.south west)+(2.5,1)$)
{
a_{1,1,3} & a_{1,2,3} & a_{1,3,3} & a_{1,4,3} \\
a_{2,1,3} & a_{2,2,3} & a_{2,3,3} & a_{2,4,3} \\
a_{3,1,3} & a_{3,2,3} & a_{3,3,3} & a_{3,4,3} \\
a_{4,1,3} & a_{4,2,3} & a_{4,3,3} & a_{4,4,3} \\
};

\matrix (mC) [draw,matrix of math nodes] at ($(mB.south west)+(2.5,1)$)
{
a_{1,1,2} & a_{1,2,2} & a_{1,3,2} & a_{1,4,2} \\
a_{2,1,2} & a_{2,2,2} & a_{2,3,2} & a_{2,4,2} \\
a_{3,1,2} & a_{3,2,2} & a_{3,3,2} & a_{3,4,2} \\
a_{4,1,2} & a_{4,2,2} & a_{4,3,2} & a_{4,4,2} \\
};

\matrix (mD) [draw,matrix of math nodes] at ($(mC.south west)+(2.5,1)$)
{
a_{1,1,1} & a_{1,2,1} & a_{1,3,1} & a_{1,4,1} \\
a_{2,1,1} & a_{2,2,1} & a_{2,3,1} & a_{2,4,1} \\
a_{3,1,1} & a_{3,2,1} & a_{3,3,1} & a_{3,4,1} \\
a_{4,1,1} & a_{4,2,1} & a_{4,3,1} & a_{4,4,1} \\
};

\draw[dashed](mA.north east)--(mD.north east);
\draw[dashed](mA.north west)--(mD.north west);
\draw[dashed](mA.south east)--(mD.south east);
\end{tikzpicture}

	\begin{gather*}
	a_{ijk}=P(\Gamma_{r}=\gamma_{k}/\Gamma_{r-1}=\gamma_{j},\Gamma_{r-2}=\gamma_{i})
	\end{gather*}
	\caption{Transition matrix used in full second order HMM}
\label{fig:fig2}
	\end{figure}
	\end{tcolorbox}

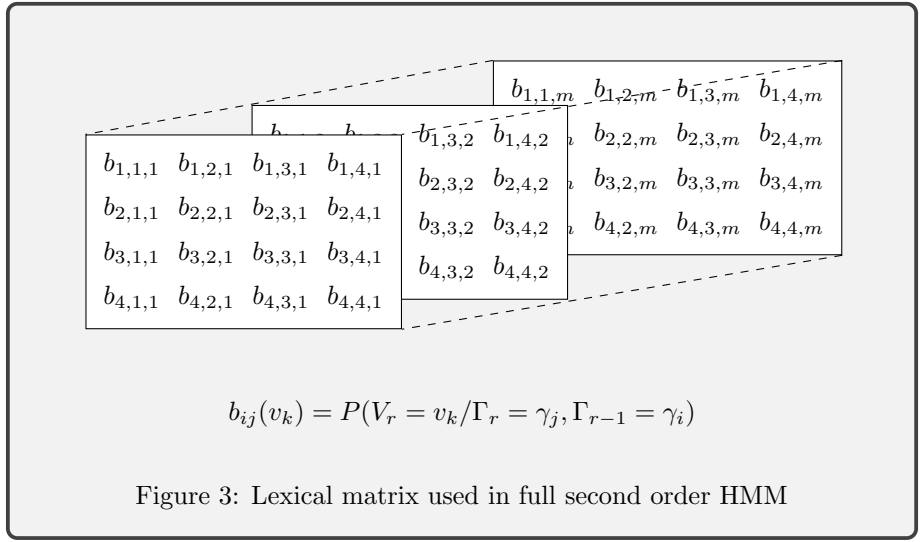
\begin{tcolorbox}
	\begin{figure}[H]
	\begin{tikzpicture}[every node/.style={anchor=north east,fill=white,minimum width=1mm,minimum height=1mm}]
\hspace*{0.5cm}
\matrix (mA) [draw,matrix of math nodes]
{
b_{1,1,m} & b_{1,2,m} & b_{1,3,m} & b_{1,4,m} \\
b_{2,1,m} & b_{2,2,m} & b_{2,3,m} & b_{2,4,m} \\
b_{3,1,m} & b_{3,2,m} & b_{3,3,m} & b_{3,4,m} \\
b_{4,1,m} & b_{4,2,m} & b_{4,3,m} & b_{4,4,m} \\
};

\matrix (mC) [draw,matrix of math nodes] at ($(mA.south west)+(1,2)$)
{
b_{1,1,2} & b_{1,2,2} & b_{1,3,2} & b_{1,4,2} \\
b_{2,1,2} & b_{2,2,2} & b_{2,3,2} & b_{2,4,2} \\
b_{3,1,2} & b_{3,2,2} & b_{3,3,2} & b_{3,4,2} \\
b_{4,1,2} & b_{4,2,2} & b_{4,3,2} & b_{4,4,2} \\
};

\matrix (mD) [draw,matrix of math nodes] at ($(mC.south west)+(2,2.2)$)
{
b_{1,1,1} & b_{1,2,1} & b_{1,3,1} & b_{1,4,1} \\
b_{2,1,1} & b_{2,2,1} & b_{2,3,1} & b_{2,4,1} \\
b_{3,1,1} & b_{3,2,1} & b_{3,3,1} & b_{3,4,1} \\
b_{4,1,1} & b_{4,2,1} & b_{4,3,1} & b_{4,4,1} \\
};

\draw[dashed](mA.north east)--(mD.north east);
\draw[dashed](mA.north west)--(mD.north west);
\draw[dashed](mA.south east)--(mD.south east);
\end{tikzpicture}

	\begin{gather*}
	b_{ij}(v_k)=P(V_{r}=v_{k}/\Gamma_{r}=\gamma_{j},\Gamma_{r-1}=\gamma_{i})
	\end{gather*}
	\caption{Lexical matrix used in full second order HMM}
\label{fig:fig3}
	\end{figure}
\end{tcolorbox}

\subsubsection{Prediction with the introduction of collected Grammatical Features}

At this point of the research, grammatical features are still not introduced on the model. When we observed the elements constituting the full second-order Hidden Markov Model contextual probabilities and lexical probabilities are formed to include second order information. What we tried to do on this stage is to keep the trigram property of the model but modifying the parameters of the contextual and lexical probabilities to introduce collected grammatical features. The changes made are: the trigram transition probability $a_{ijk}=P(\Gamma_{r}=\gamma_{k}/\Gamma_{r-1}=\gamma_{j},\Gamma_{r-2}=\gamma_{i})$ is replaced by the probability $a_{ijk}=P(\Gamma_{r}=\gamma_{k}/\Gamma_{r-1}=\gamma_{j},T_{r-1}=\tau_{i})$ and the trigram lexical probability $b_{ij}(v_k)=P(V_{r}=v_{k}/\Gamma_{r}=\gamma_{j},\Gamma_{r-1}=\gamma_{i})$ is replaced by the probability $b_{ij}(v_k)=P(V_{r}=v_{k}/\Gamma_{r}=\gamma_{j},T_{r}=\tau_{i})$.The elements of the HMM becomes:
\begin{itemize}
\item N=4, the number of distinct elements of the set $\Gamma$.
\item M=809, number of tokens.
\item K=14, the number of distinct POS tags.
\item Matrix A containing the state transition probability distribution. This matrix is of dimensions  $N \mtimes N \mtimes K$ as shown on {\color{blue}figure \ref{fig:fig4}}.

\item Matrix B containing the observation symbol probability distribution. This matrix is of dimensions  $M \mtimes N \mtimes K$ as shown on {\color{blue}figure \ref{fig:fig5}}.

\item $\pi$ a vector containing the probability distributions that a sentence begins with one of the elements of $\Gamma$
\begin{gather*}
	\pi=\begin{bmatrix}
	\pi_{1} & \pi_{2} & \pi_{3} & \pi_{4} 
	\end{bmatrix}
	\end{gather*}
\end{itemize}

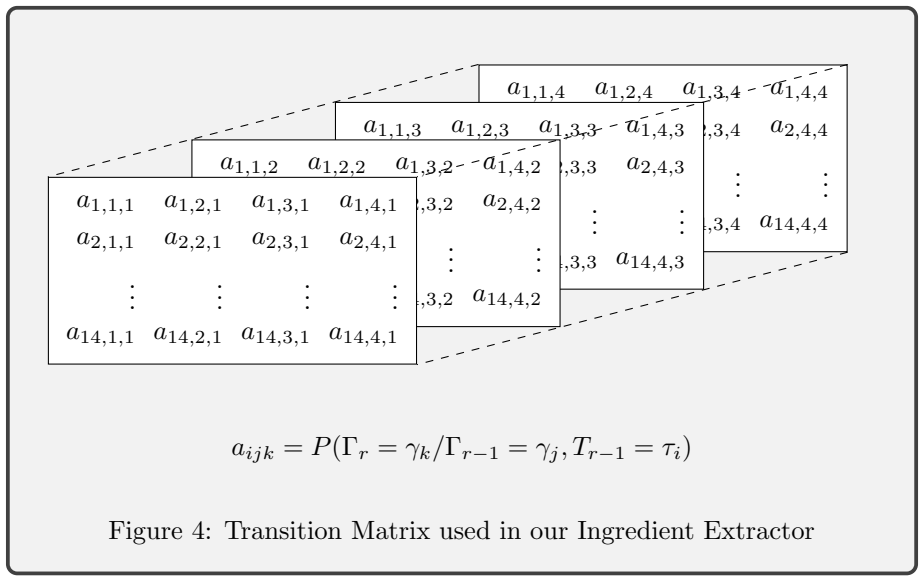
\begin{tcolorbox}
\begin{figure}[H]
\begin{tikzpicture}[every node/.style={anchor=north east,fill=white,minimum width=1mm,minimum height=2mm}]
\hspace*{-0cm}
\matrix (mA) [draw,matrix of math nodes]
{
a_{1,1,4} & a_{1,2,4} & a_{1,3,4} & a_{1,4,4} \\
a_{2,1,4} & a_{2,2,4} & a_{2,3,4} & a_{2,4,4} \\
\vdots & \vdots & \vdots & \vdots \\
a_{14,1,4} & a_{14,2,4} & a_{14,3,4} & a_{14,4,4} \\
};

\matrix (mB) [draw,matrix of math nodes] at ($(mA.south west)+(3,2)$)
{
a_{1,1,3} & a_{1,2,3} & a_{1,3,3} & a_{1,4,3} \\
a_{2,1,3} & a_{2,2,3} & a_{2,3,3} & a_{2,4,3} \\
\vdots & \vdots & \vdots & \vdots \\
a_{14,1,3} & a_{14,2,3} & a_{14,3,3} & a_{14,4,3} \\
};

\matrix (mC) [draw,matrix of math nodes] at ($(mB.south west)+(3,2)$)
{
a_{1,1,2} & a_{1,2,2} & a_{1,3,2} & a_{1,4,2} \\
a_{2,1,2} & a_{2,2,2} & a_{2,3,2} & a_{2,4,2} \\
\vdots & \vdots & \vdots & \vdots \\
a_{14,1,2} & a_{14,2,2} & a_{14,3,2} & a_{14,4,2} \\
};

\matrix (mD) [draw,matrix of math nodes] at ($(mC.south west)+(3,2)$)
{
a_{1,1,1} & a_{1,2,1} & a_{1,3,1} & a_{1,4,1} \\
a_{2,1,1} & a_{2,2,1} & a_{2,3,1} & a_{2,4,1} \\
\vdots & \vdots & \vdots & \vdots \\
a_{14,1,1} & a_{14,2,1} & a_{14,3,1} & a_{14,4,1} \\
};

\draw[dashed](mA.north east)--(mD.north east);
\draw[dashed](mA.north west)--(mD.north west);
\draw[dashed](mA.south east)--(mD.south east);
\end{tikzpicture}

	\begin{gather*}
	a_{ijk}=P(\Gamma_{r}=\gamma_{k}/\Gamma_{r-1}=\gamma_{j},T_{r-1}=\tau_{i})
	\end{gather*}
	\caption{Transition Matrix used in our Ingredient Extractor}
\label{fig:fig4}
	\end{figure}
	\end{tcolorbox}
	
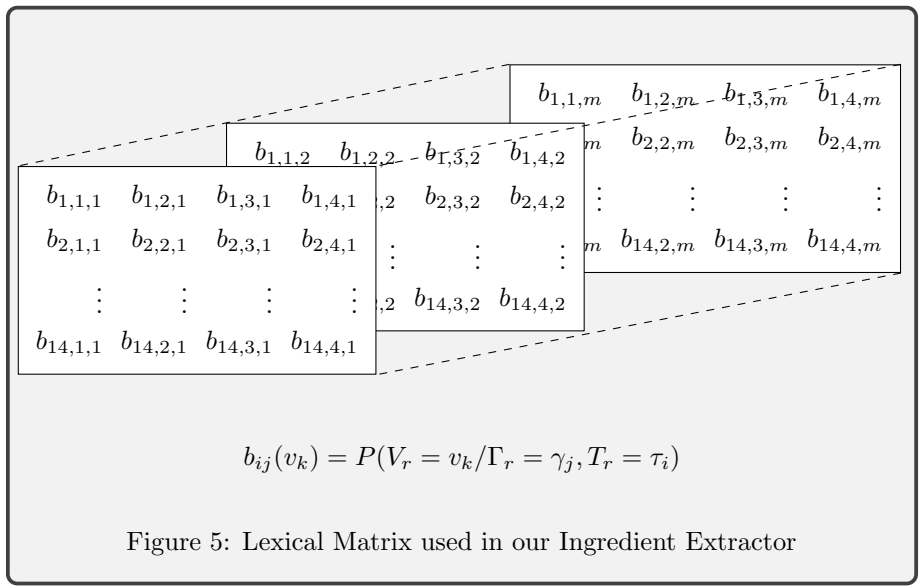
\begin{tcolorbox}
	\begin{figure}[H]
	\begin{tikzpicture}[every node/.style={anchor=north east,fill=white,minimum width=1mm,minimum height=1mm}]
\hspace*{-0.4cm}
\matrix (mA) [draw,matrix of math nodes]
{
b_{1,1,m} & b_{1,2,m} & b_{1,3,m} & b_{1,4,m} \\
b_{2,1,m} & b_{2,2,m} & b_{2,3,m} & b_{2,4,m} \\
\vdots & \vdots & \vdots & \vdots \\
b_{14,1,m} & b_{14,2,m} & b_{14,3,m} & b_{14,4,m} \\
};

\matrix (mC) [draw,matrix of math nodes] at ($(mA.south west)+(1,2)$)
{
b_{1,1,2} & b_{1,2,2} & b_{1,3,2} & b_{1,4,2} \\
b_{2,1,2} & b_{2,2,2} & b_{2,3,2} & b_{2,4,2} \\
\vdots & \vdots & \vdots & \vdots \\
b_{14,1,2} & b_{14,2,2} & b_{14,3,2} & b_{14,4,2} \\
};

\matrix (mD) [draw,matrix of math nodes] at ($(mC.south west)+(2,2.2)$)
{
b_{1,1,1} & b_{1,2,1} & b_{1,3,1} & b_{1,4,1} \\
b_{2,1,1} & b_{2,2,1} & b_{2,3,1} & b_{2,4,1} \\
\vdots & \vdots & \vdots & \vdots \\
b_{14,1,1} & b_{14,2,1} & b_{14,3,1} & b_{14,4,1} \\
};

\draw[dashed](mA.north east)--(mD.north east);
\draw[dashed](mA.north west)--(mD.north west);
\draw[dashed](mA.south east)--(mD.south east);
\end{tikzpicture}

	\begin{gather*}
	b_{ij}(v_k)=P(V_{r}=v_{k}/\Gamma_{r}=\gamma_{j},T_{r}=\tau_{i})
	\end{gather*}
	\caption{Lexical Matrix used in our Ingredient Extractor}
\label{fig:fig5}
	\end{figure}
\end{tcolorbox}

\section{Calculating Probabilities for Unknown Words }
One common problem in part of speech tagging processing is to predict the hidden state of an unknown word. 

Andrei Mikheev{\color{blue}\cite{mikheev1996learning}} constructed an automatic technique for learning English part-of-speech guessing rules. 

Scott M.Thede{\color{blue}\cite{thede1999second}} used a Second-Order Hidden Markov Model for POS tagging by estimating the probability $P( $word has suffix $s_{k}/$tag is $ t_{j})$ 

when he tried to predict the hidden state of an unknown word. 

Scott M.Thede{\color{blue}\cite{thede1999second}} used a trigram tagger respecting the elements that constitute an HMM explained in details by Lawrence R. Rabiner {\color{blue}\cite{rabiner1989tutorial}} 

On this work we tried to create a first-order Hidden Markov Model POS Arabic Tagger and we choose the prefix of the word as a word feature rather than word suffix as described by Thede.  

After training the corpus we can estimate the probability $P(v_{i}/t_{i})$ by counting the number of times the hidden state $t_{i}$ was attributed to the word $v_{i}$. 

For example, we had a corpus constituted of 4 words as represented on the {\color{blue}Table \ref{tab:tab3}}. We trained the corpus with 3 hidden states and we obtain the discrete state-dependent word probabilities table: 

\begin{table}[H]
\begin{center}
		\begin{tabular}{|l|c|c|c|r|}
			\cline{2-5}
			\multicolumn{1}{c|}{} & 2 & \AR{البرتقال}  & \AR{الاكليل} & \AR{أو}  \\
			\hline
				B & 1 & 0 & 0 & 0 \\ \hline
				C & 0 & 0.5 & 0.5 & 0 \\
			\hline
			M & 0 & 0 & 0  & 1 \\
			\hline
			
		\end{tabular}
		\caption{The discrete state-dependent word probabilities table for a mini-corpus}
		\label{tab:tab3}
		\end{center}
		\end{table}
		
For example, the probability $P(w_{i}=$\AR{البرتقال}$/t_{i}=C)$ is estimated by counting the number of times the word \AR{البرتقال} appears with the tag C divided by the number of times C appears on the corpus. 

In Arabic as in English, words are constituted morphologically based on its grammatical role on a sentence. This is why suffix and prefix of a word can help us to predict its tag. 

Andrei Mikheev {\color{blue}\cite{mikheev1996learning}} presented a list of morphological rules for English to predict unknown words. For example, the word we obtain by adding the prefix "un" to a word that was tagged with the tags (VBD) and (VBN) can play the role of an adjective on a sentence.  

Such morphological rule can be also applied to Arabic Language. For example, adding "\AR{ال}" to a word that played the role of a noun "\AR{إسم}" or a noun that comes before the adjective "\AR{منعوت}" can be estimated as a Known Noun "\AR{إسم معرف}". 

Scott M.Thede {\color{blue}(Thede,1998)} was inspired by morphological rules described by Mikheev to construct his unknown word predictor. He first created a lexicon to estimate $P(w_{i}/t_{i})$ , then he used prefix and suffix of words on this lexicon to predict possible tags. This method can be used to our corpus as in Arabic word feature extraction can be done by the prefix of the words. 

For example, the discrete state-dependent word probabilities table for the previous mini-corpus can be represented depending on word prefixes: 
\begin{table}[H]
		\begin{tabular}{|l|c|c|c|r|}
			\cline{2-5}
			\multicolumn{1}{c|}{} & 2 & \AR{ال}  & \AR{ال} & \AR{أو}  \\
			\hline
				B & 1 & 0 & 0 & 0 \\ \hline
				C & 0 & 0.5 & 0.5 & 0 \\
			\hline
			M & 0 & 0 & 0  & 1 \\
			\hline
		\end{tabular}
		\centering
\caption{The discrete state-dependent word probabilities table for a mini-corpus (prefix-based approach)}
\label{tab:tab4}
\end{table}
We considered the two first characters of a word on this work if a word counts more than two characters and the first character of a word if the word is constituted with only one character.
After that a probability distribution for each affix is created by adding the probabilities $P(v_{i}/t_{i})$ for words with the same tag. For example, $P(prefix=$"\AR{ال}"$/t_{i}=C)=P(v_{i}=$\AR{البرتقال}$/t_{i}=C)+P(v_{i}=$\AR{الاكليل}$/t_{i}=C)$ the {\color{blue}Table \ref{tab:tab4}} becomes:
\begin{table}[H]
		\begin{tabular}{|l|c|c|r|}
			\cline{2-4}
			\multicolumn{1}{c|}{} & 2  & \AR{ال} & \AR{أو}  \\
			\hline
				B & 1 & 0 & 0 \\ \hline
				C & 0 & 1 & 0 \\
			\hline
			M & 0 & 0 & 1 \\
			\hline
		\end{tabular}
		\centering
\caption{The discrete state-dependent word probabilities table for a mini-corpus  after adding the probabilities $P(v_{i}/t_{i})$ for words with the same tag (prefix-based approach)}
\label{tab:tab5}
	\end{table}
{\color{blue}Table \ref{tab:tab5}} constitute the matrix C used on the Hidden Markov Model algorithm to replace the matrix B when a word is unknown.

\begin{table}[h!]
\begin{center}
		\begin{tabular}{|l|c|r|}
			\cline{2-3}
			\multicolumn{1}{c|}{} & Accuracy  & F1 score   \\
			\hline
				\makecell{First order Hidden \\ Markov Model applied to tokens \\ to predict ingredients} & 96.64 \% & 70.07 \% \\ \hline
				\hline
				\makecell{Second order Hidden \\ Markov Model applied to tokens \\ to predict ingredients} & 95.85 \% & 68.22 \% \\ \hline
			\hline
				\makecell{First Order Hidden Markov Model \\ applied to tags \\ to predict ingredients} & 79.35 \% & 54.36 \%  \\ \hline
				\hline
				\makecell{Second Order Hidden Markov Model \\ applied to tags \\ to predict ingredients} & 87.74 \% & 58.53 \%  \\ \hline
				\hline
				\makecell{Our ingredient extractor \\ with 100 \% accuracy on the first layer \\ with $\lambda=4$} & 98.44 \% & 81.44 \%  \\ \hline \hline
				\makecell{Our ingredient extractor \\ with 90.33 \% accuracy on the first layer \\ with 82.03 \% F1 score on the first layer \\ with $\lambda=4$} & 97.08 \% & 74.36 \%  \\ \hline
		\end{tabular}
		\centering
\caption{ Comparison of performances between our ingredient extractor and other HMM taggers using the trained corpus as a testing dataset to avoid unknown words}
\label{tab:tab6}
\end{center}
\end{table}

	\begin{figure}[h!]
\begin{center}
\begin{tikzpicture}
\begin{axis}[
grid=both,
ticks=both,
legend pos=outer north east]
\draw [red, thick] ({rel axis cs:0,0}-|{axis cs:4,0}) -- ({rel axis cs:0,1}-|{axis cs:4,0}) node  [pos=0.25] {$\lambda_{max}=4$};
\addplot[red] coordinates {
(1,0.917)(2,0.964)(3,0.9805)(4,0.983)(5,0.98)(6,0.982)(7,0.982)(8,0.983)(9,0.983)
};
\addplot[blue] coordinates {
(1,0.915)(2,0.954)(3,0.968)(4,0.97)(5,0.968)(6,0.968)(7,0.967)(8,0.966)(9,0.966)
};
\addplot[green] coordinates {
(1,0.8953)(2,0.9372)(3,0.9455)(4,0.9454)(5,0.9418)(6,0.941)(7,0.9364)(8,0.9343)(9,0.9307)
};
\legend{100 \% accuracy  ,90.3 \% accuracy, 67.8 \% accuracy}
\end{axis}
\end{tikzpicture}
\caption{ Variations of accuracy of our Ingredient Extractor depending on $\lambda$ and accuracy score on the first layer}
\label{fig6}
\end{center}
\end{figure}
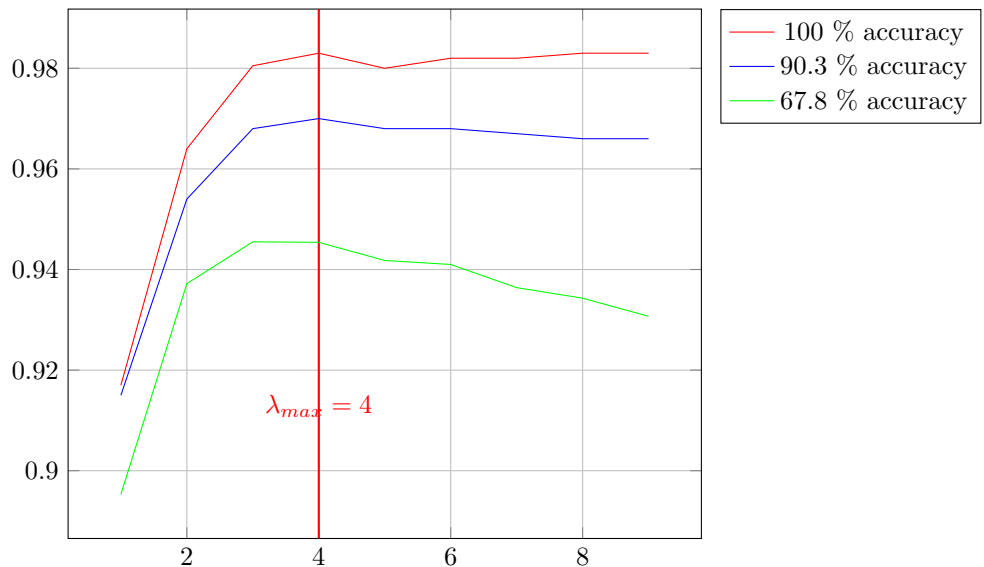

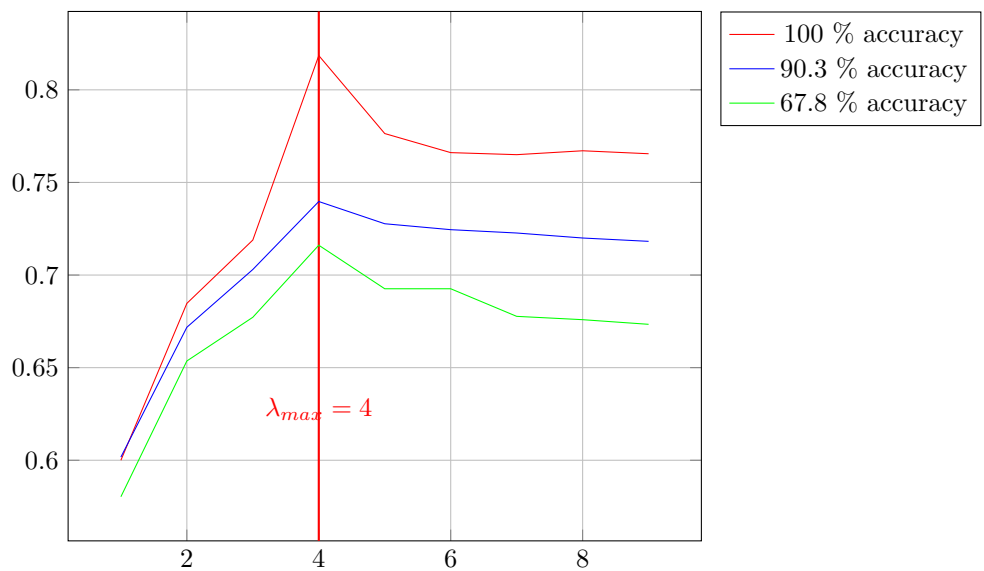
\begin{figure}[h!]
\begin{center}
\begin{tikzpicture}
\begin{axis}[
grid=both,
ticks=both,
legend pos=outer north east]
\draw [red, thick] ({rel axis cs:0,0}-|{axis cs:4,0}) -- ({rel axis cs:0,1}-|{axis cs:4,0})  node  [pos=0.25] {$\lambda_{max}=4$};
\addplot[red] coordinates {
(1,0.6)(2,0.6847)(3,0.7189)(4,0.8185)(5,0.7764)(6,0.7661)(7,0.765)(8,0.7671)(9,0.7655)
};
\addplot[blue] coordinates {
(1,0.6017)(2,0.6718)(3,0.703)(4,0.7397)(5,0.7277)(6,0.7245)(7,0.7227)(8,0.72)(9,0.7182)
};
\addplot[green] coordinates {
(1,0.5803)(2,0.6536)(3,0.6772)(4,0.7161)(5,0.6926)(6,0.6926)(7,0.6777)(8,0.6759)(9,0.6734)
};
\legend{100 \% accuracy  ,90.3 \% accuracy, 67.8 \% accuracy}
\end{axis}
\end{tikzpicture}
\caption{Variations of F1 score of our Ingredient Extractor depending on $\lambda$ and accuracy score on the first layer}
\label{fig7}
\end{center}
\end{figure}

\begin{table}[p]
\begin{center}
    \begin{tabular}{| l | l | l | l | l | l | l |}
    \hline
    \makecell{tag} & accuracy & f1 score & \makecell{accuracy for \\ unknown \\ words} & \makecell{accuracy for \\ known \\ words} & \makecell{number of \\ unknown \\ words} & \makecell{unknown \\ words \\ percentage} \\ \hline
   0& 94.77 \% & 67.67 \% & 67.78 \% & 96.33 \% & 78 & 9.67 \%  \\ \hline
    1 & 93.02 \% & 65.56 \% & 63.93 \% & 95.2 \% & 107 & 13.26 \%  \\ \hline
    2 &  94.15 \% & 67.44 \% & 69.9 \% & 95.81 \% & 88& 10.9 \%  \\
    \hline
    3  & 94.01 \% & 87.89 \% & 62.65 \% & 95.7 \% & 75 & 9.29 \%  \\ \hline
   4  & 93.14 \% & 87.59 \% & 67.96 \% & 94.73 \% & 92 & 11.4 \% \\ \hline
   5&  95.78 \% & 69.61 \% & 76.4 \% & 96.83 \% & 83 & 10.29 \% \\
    \hline 
    6&  94.38 \% & 67.53 \% & 54.65 \% & 96.49 \% & 80 & 9.91 \%  \\ \hline
    7& 95.08 \% & 91.17 \% & 61.22 \% & 97.17 \% & 90 & 11.15 \%  \\ \hline
    8&  96.83 \% & 94.04 \% & 79.76 \% & 97.7 \% & 77 & 9.54 \% \\ \hline
    9 & 94.92 \% & 90.7 \% & 64.95 \% & 96.72 \% & 88 & 10.9 \%  \\ \hline \hline
   Avg & 94.61 \% & 78.92 \% & 66.92 \% & 96.27 \% & 86 & 10.63 \% \\ \hline
  
    \end{tabular}
    \label{tab:tab10}
\end{center}
\caption{Average performances of first order HMM used to predict ingredients state using 10 fold cross-validation and a 80\% training dataset and 20 \% testing dataset separated}
\end{table}

\begin{table}[p]
\begin{center}
    \begin{tabular}{| l | l | l | l | l | l | l |}
    \hline
    \makecell{tag} & accuracy & f1 score & \makecell{accuracy for \\ unknown \\ words} & \makecell{accuracy for \\ known \\ words} & \makecell{number of \\ unknown \\ words} & \makecell{unknown \\ words \\ percentage} \\ \hline
   0& 93.72 \% & 64.13 \% & 70.0 \% &94.81 \% & 78 & 9.67 \%  \\ \hline
    1 &92.53 \% & 63 \% & 64.75 \% &94.19 \% & 107 & 13.26 \%  \\ \hline
    2 &94.14 \% & 65.6 \% & 70.87 \% &95.39 \% & 88 & 10.9 \%  \\
    \hline
    3  &93.94 \% & 86.92 \% & 68.67 \% &95.02\% & 75 & 9.29 \%   \\ \hline
   4  &93.42 \% & 86.61 \% & 66.02 \% &94.81 \% & 92 & 11.4 \%  \\ \hline
   5&93.83 \% & 65.85 \% & 62.92 \% &95.18 \% & 83 & 10.29 \% \\
    \hline 
    6& 93.76 \% & 65.63 \% & 60.47 \% &95.19 \% & 80 & 9.91 \%  \\ \hline
    7& 94.04 \% & 88.56 \% & 55.1 \% &95.96 \% & 90 & 11.15 \% \\ \hline
    8& 95.34 \% & 90.99 \% & 60.71 \% &96.77 \% & 77 & 9.54 \%  \\ \hline
    9 & 94.34 \% & 88.56 \% & 61.86 \% &95.91 \% & 88 & 10.9 \%  \\ \hline \hline
   Avg & 93.91 \% & 76.58 \% & 64.14 \% &95.32 \% & 86 & 10.63 \% \\ \hline
  
    \end{tabular}
    \label{tab:tab11}
\end{center}
\caption{Average performances of second order HMM used to predict ingredients state using 10 fold cross-validation and a 80\% training dataset and 20 \% testing dataset separated }
\end{table}

\begin{table}[p]
\begin{center}
    \begin{tabular}{| l | l | l | l | l | l | l |}
    \hline
    \makecell{tag} & accuracy & f1 score & \makecell{accuracy for \\ unknown \\ words} & \makecell{accuracy for \\ known \\ words} & \makecell{number of \\ unknown \\ words} & \makecell{unknown \\ words \\ percentage} \\ \hline
   0& 95.78 \% & 67.91 \% & 61.11 \% & 97.38 \% & 78 & 9.67 \%  \\ \hline
    1 & 94.38  \% & 66.54 \% & 68.85 \% & 95.92 \% & 107 & 13.26 \%  \\ \hline
    2 & 95.08  \% & 67.39 \% & 67.96 \% & 96.54 \% & 88  & 10.9 \%  \\
    \hline
    3  & 95.83  \% & 67.71 \% & 62.65 \% & 97.25 \% & 75  & 9.29 \%  \\ \hline
   4  & 94.83  \% & 67.72 \% & 71.84 \% & 96 \% & 92  & 11.4 \%  \\ \hline
   5& 95.9  \% & 69.57 \% & 70.79 \% & 97.0 \% & 83  & 10.29 \%  \\
    \hline 
    6& 94.43  \% & 65.92 \% & 55.81 \% & 96.08 \% & 80  & 9.91 \%  \\ \hline
    7&94.13  \% & 65.61 \% & 52.04 \% & 96.22 \% & 90  & 11.15 \%  \\ \hline
    8& 95.63  \% & 68.82 \% & 60.71 \% & 97.06 \% & 77  & 9.54 \% \\ \hline
    9 &94.39  \% & 65.95 \% &56.7 \% & 96.21 \% &88  & 10.9\%  \\ \hline \hline
   Avg & 95.04  \% & 67.31 \% &62.85 \% & 96.57 \% &86  & 10.63\% \\ \hline
  
    \end{tabular}
    \label{tab:tab12}
\end{center}
\caption{Average performances in predicting ingredients states with $\lambda=4$ when accuracy is 100 \% on first layer using 10 fold cross-validation and a 80\% training dataset and 20\% testing dataset separated  }
\end{table}

\begin{table}[p]
\begin{center}
    \begin{tabular}{| l | l | l | l | l | l | l | l | l |}
    \cline{2-9}
		\multicolumn{1}{c|}{}  &	\multicolumn{4}{c|}{\makecell{First Layer \\ first order \\ HMM}} & \multicolumn{4}{c|}{\makecell{Second Layer \\ second order \\ HMM}}\\
    \hline
    Fold & accuracy & f1 score & \makecell{accuracy \\ unknown \\ words} & \makecell{accuracy \\ known \\ words} & accuracy & f1 score & \makecell{ accuracy \\ unknown \\ words } & \makecell{accuracy \\ known \\ words}  \\ \hline
    0 &90.33 \% &70.61 \% & 50.55 \% & 92.14 \% & 95.43 \% & 66.98 \% & 61.54 \% & 96.97 \% \\ \hline
    1 & 87.7 \% & 73.71 \% & 52.03 \% & 89.81 \% & 93.96 \% &65.96 \% & 70.73 \% & 95.32 \% \\ \hline
    2 & 89.07 \% & 75.34 \% &66.99 \% & 90.26 \% &94.54 \% &66.97 \% & 66.02 \% & 96.07 \% \\ \hline
    3 & 87.87 \% & 72.36 \%  & 67.47 \% & 88.75 \% & 94.93 \% & 66.93 \% & 63.86 \% & 96.27 \% \\ \hline
    4 & 87.31 \% &73.21 \% & 63.11 \% & 88.54 \% & 93.65 \% & 66.46 \% & 72.82 \% &94.71 \% \\ \hline
    5 & 91.23 \% & 76.51 \% & 62.92 \% & 92.47 \% & 94.82 \% & 68.17 \% & 69.66 \% & 95.92 \% \\
    \hline
   6 & 90.53 \% & 73.87 \% & 59.3 \% & 91.86 \% & 93.86 \% & 65.16 \% & 54.65 \% & 95.92 \% \\
    \hline
    7 & 88.13 \% & 78.5 \% & 61.22 \% & 90.51 \% & 93.8 \% & 65.12 \% & 55.1 \% & 95.71 \% \\
    \hline
    8 & 89.28 \% & 77.76 \% & 67.86 \% & 90.16 \% & 95.01 \% & 67.84 \% & 64.29 \% & 96.28 \% \\ \hline
    9 & 89.4 \% & 77.29 \% & 65.98 \% & 90.53 \% & 93.96 \% & 64.98 \% & 60.82 \% & 95.57 \% \\
    \hline \hline
    Avg & 89.19 \% & 74.92 \% & 61.74 \% &90.5 \% & 94.4 \% & 66.46 \% & 63.95 \% & 95.83 \% \\
    \hline
    
    \end{tabular}
\end{center}
\caption{Average performances in predicting ingredients states with various accuracies in the first layer with $\lambda=4$ when accuracy is 100 \% on first layer using 10 fold cross-validation and a 80\% training dataset and 20\% testing dataset separated t}
\label{tab13}
\end{table}

	\begin{center}

\begin{table}[h!]
		\begin{tabular}{|l|c|c|c|r|}
			\cline{2-5}
			\multicolumn{1}{c|}{} & known  & unknown & overall & F1 score \\
			\hline
				\makecell{First order Hidden \\ Markov Model applied to tokens} & 96.27 \% & 66.92 \% & 94.61 \% & 78.92 \% \\ \hline
				\hline
				\makecell{Second order Hidden \\ Markov Model applied to tokens} & 95.32 \% & 64.14 \% & 93.91 \% & 76.58 \% \\ \hline
			\hline
				\makecell{Our ingredient extractor \\ with 100 \% accuracy on the first layer \\ with $\lambda=4$} & 96.57 \% & 62.85 \% & 95.02 \% & 67.31 \% \\ \hline
				\hline
				\makecell{Our ingredient extractor \\ with 89.19 \% average accuracy on the first layer \\ with 74.92 \% average  F1 score on the first layer \\ with $\lambda=4$} & 95.83 \% & 63.95 \% & 94.4 \% & 66.46 \% \\ \hline
			
		\end{tabular}

		\centering
\caption{Comparison of performances between the 3 methods }
\label{jkl}
\end{table}
	\end{center}

\newpage

\section{Results and Interpretation}

First, we tested the Modified Viterbi algorithm used in our Ingredient Extractor on the same trained corpus to avoid using matrix C described in {\color{blue}Table (\ref{tab:tab5}}) . Our Ingredient Extractor shows the highest accuracy and F1 score even when accuracy on the first layer don't overcome 91\%  {\color{blue}Table (\ref{tab:tab6}}). These first results before dividing the dataset into two separate training and testing sides are encouraging and shows that including POS tags inside the calculation of a state matrix isn't a bad idea because the state of an ingredient depend also from the state of the POS tag.

Second, we tested the Modified Viterbi algorithm used in our Ingredient Extractor on a testing dataset different from the trained dataset and we compare the results obtained with the results for a $1^{st}$ order HMM and a $2^{nd}$ order HMM using 10-fold cross-validation. We obtain in {\color{blue}Table (\ref{jkl}}) an f1 score of 67.31 \% for our IE when the first layer has a 100 \% accuracy against 78.92 \% F1 score for a $1^{st}$ order HMM and 76.58 \% F1 score for a $2^{nd}$ order HMM, our F1 score decreased with a percentage of 14.13 \% when we split the dataset into a training and a testing dataset for our IE. This decrease on F1 score could be explained by the change made in the lexical probability when it is confronted with unknown words making its contribution in the state matrix not as efficient as when the entire word is considered in the lexical probability. The hyper parameter $\lambda$ is eliminated from our IE when the word to be estimated is unknown for a better accuracy and F1 score.

\section{Conclusion}

Our Ingredient Extractor algorithm showed great results. It is based on HMM methods. We realized it by training two layers: first we trained tokens by tagging POS tags and second, we trained tokens by extracting the ingredients. Our HMM model needed modifications in iteration step because we didn't get a square transition probability matrix or a square lexical probability matrix after the training step. A detailed iterations of our method is illustrated in appendix \ref{sec:num1}. We can ameliorate our model by calculating the probabilities in it differently not as simple as we deed. We can make our model more interesting by adding two layers, one for extracting quantities and the other for extracting unities. 

\newpage
\appendix
\section{Detailed Iterations of Modified Second Order Viterbi Algorithm Used in our Ingredient Extractor} \label{sec:num1}
\begin{enumerate}
	\item Predicting Tags for the first Layer:

	We use for predicting tags a first order hidden Markov model as this method have a better accuracy than second order hidden Markov model.
	
	\item Predicting Ingredients state for the second Layer:
	
	 \textbf{The variables:}
	 \begin{itemize}
		\item $\delta_{l}(i,j)=max_{\tau_{1}...\tau_{l-1}}P(\tau_{1}...\tau_{l}=[t_{i},\gamma_{j}]/v_{1}...v_{l},t_{1}...t_{l}), l={1...L}$
		\item $\psi_{l}(i,j)=argmax_{\tau_{1}...\tau_{l-1}}P(\tau_{1}...\tau_{l}=[t_{i},\gamma_{j}]/v_{1}...v_{l},t_{1}...t_{l}), l={1...L}$
	\end{itemize}
	\textbf{The procedure:}
	\begin{enumerate}
		\item Initialization step:
		\begin{gather*} 
		\delta_{1}(i,j) = \begin{cases}
	\pi_{j}b_{ij}(v_{1}) & v_{1} \text{ is known} \text{ i={1..14} j={1..4} } \\
	\,  \\
	\pi_{j}c_{ij}(v_{1}) & v_{1} \text{ is unknown} \text{ i={1..14} j={1..4} } \\
	\end{cases}\\
		 \end{gather*}
		 \begin{gather*}
	\psi_{1}(i,j)=0  \text{ i={1..14} j={1..4}}
	\end{gather*}
		\item Iteration step:
		\begin{gather*} 
		\delta_{l}(i,j) = \begin{cases}
	max_{j}[\delta_{l-1}(i,j)a_{ijk}]b_{ij}(v_{l}) & v_{l} \text{ is known , l={2...L} } \\ & \text{length  j={1...4}, i={1..14}} \\ & \text{k={1...4}}\\
	\,  \\
	max_{j}[\delta_{l-1}(i,j)a_{ijk}]c_{ij}(v_{l}) & v_{l} \text{ is unknown , l={2...L} } \\ & \text{ j={1...4}, i={1..14}} \\ & \text{k={1...4}}
	\end{cases}\\
		 \end{gather*}
		  \begin{gather*}
	\psi_{l}(i,j)=argmax_{j}[\delta_{l-1}(i,j)a_{ijk}]  \text{j={1..4},i={1..14},k={1..4}}
	\end{gather*}
	\item Termination:
		 \begin{gather*} 
\tau_{L}^{*}=argmax_{i=y(L),j=1..4}\delta_{L}(i,j)   
\\
\text{ where y(L) is the tag calculated at previous layer in position L}
		 \end{gather*}
		 \item Backtracking:
		\begin{gather*} 
		\tau_{l}^{*}=\psi_{l}(y(l),\tau_{l+1}^{*})   \text{ l=L-1..2,1} 
		\\
\text{ where y(l) is the tag calculated at previous layer in position l}
		 \end{gather*}
		 
	\end{enumerate}
	\item Predicting Ingredients state for the second Layer in Log-space and introduction of the hyper-parameter $\lambda$
	
	\textbf{The variables:}
	 \begin{itemize}
		\item $\delta_{l}(i,j)=max_{\tau_{1}...\tau_{l-1}}P(\tau_{1}...\tau_{l}=[t_{i},\gamma_{j}]/v_{1}...v_{l},t_{1}...t_{l}), l={1...L}$
		\item $\psi_{l}(i,j)=argmax_{\tau_{1}...\tau_{l-1}}P(\tau_{1}...\tau_{l}=[t_{i},\gamma_{j}]/v_{1}...v_{l},t_{1}...t_{l}), l={1...L}$
	\end{itemize}
	\textbf{The procedure:}
	\begin{enumerate}
		\item Initialization step:
		\begin{gather*} 
		\delta_{1}(i,j) = \begin{cases}
	\log(\pi_{j}) + \log(b_{ij}(v_{1})) & v_{1} \text{ is known} \text{ i={1..14} j={1..4} } \\
	\,  \\
	\log(\pi_{j})+\log(c_{ij}(v_{1})) & v_{1} \text{ is unknown} \text{ i={1..14} j={1..4} } \\
	\end{cases}\\
		 \end{gather*}
		 \begin{gather*}
	\psi_{1}(i,j)=0  \text{ i={1..14} j={1..4}}
	\end{gather*}
		\item Iteration step:
		\begin{gather*} 
		\hspace*{-2cm}
		\delta_{l}(i,j) = \begin{cases}
	max_{j}[\delta_{l-1}(i,j)+\log(a_{ijk})]+\lambda_{max}\log(b_{ij}(v_{l})) & v_{l} \text{ is known  l={2..L} } \\ & \text{j={1...4}, i={1..14}} \\ & \text{k={1...4}}\\
	\,  \\
	max_{j}[\delta_{l-1}(i,j)+\log(a_{ijk})]+\lambda_{max}\log(c_{ij}(v_{l})) & v_{l} \text{ is unknown} \\ & \text{l={2...L} j={1..4}} \\ & \text{i={1..14} k={1..4}}
	\end{cases}\\
		 \end{gather*}
		  \begin{gather*}
	\psi_{l}(i,j)=argmax_{j}[\delta_{l-1}(i,j)+\log(a_{ijk})]  \text{ j={1..4},i={1..14},k={1..4}}
	\end{gather*}
	\item Termination:
		 \begin{gather*} 
\tau_{L}^{*}=argmax_{i=y(L),j=1..4}\delta_{L}(i,j) 
\\
\text{ where y(L) is the tag calculated at previous layer in position L}
		 \end{gather*}
		 \item Backtracking:
		\begin{gather*} 
		\tau_{l}^{*}=\psi_{l}(y(l),\tau_{l+1}^{*})   \text{ l=L-1..2,1} 
		\\
		\text{ where y(l) is the tag calculated at previous layer in position l}
		 \end{gather*}
		 
	\end{enumerate}
	\end{enumerate}

\newpage
\listoffigures
\listoftables
\bibliographystyle{apacite}
\bibliography{mybibfile}

\begin{thebibliography}{}

\bibitem [\protect \citeauthoryear {%
Ahn%
, Ahnert%
, Bagrow%
\BCBL {}\ \BBA {} Barab{\'a}si%
}{%
Ahn%
\ \protect \BOthers {.}}{%
{\protect \APACyear {2011}}%
}]{%
ahn2011flavor}
\APACinsertmetastar {%
ahn2011flavor}%
\begin{APACrefauthors}%
Ahn, Y\BHBI Y.%
, Ahnert, S\BPBI E.%
, Bagrow, J\BPBI P.%
\BCBL {}\ \BBA {} Barab{\'a}si, A\BHBI L.%
\end{APACrefauthors}%
\unskip\
\newblock
\APACrefYearMonthDay{2011}{}{}.
\newblock
{\BBOQ}\APACrefatitle {Flavor network and the principles of food pairing}
  {Flavor network and the principles of food pairing}.{\BBCQ}
\newblock
\APACjournalVolNumPages{Scientific reports}{1}{}{196}.
\PrintBackRefs{\CurrentBib}

\bibitem [\protect \citeauthoryear {%
Ahnert%
}{%
Ahnert%
}{%
{\protect \APACyear {2013}}%
}]{%
ahnert2013network}
\APACinsertmetastar {%
ahnert2013network}%
\begin{APACrefauthors}%
Ahnert, S\BPBI E.%
\end{APACrefauthors}%
\unskip\
\newblock
\APACrefYearMonthDay{2013}{}{}.
\newblock
{\BBOQ}\APACrefatitle {Network analysis and data mining in food science: the
  emergence of computational gastronomy} {Network analysis and data mining in
  food science: the emergence of computational gastronomy}.{\BBCQ}
\newblock
\APACjournalVolNumPages{Flavour}{2}{1}{1--3}.
\PrintBackRefs{\CurrentBib}

\bibitem [\protect \citeauthoryear {%
Aiello%
, Schifanella%
, Quercia%
\BCBL {}\ \BBA {} Del~Prete%
}{%
Aiello%
\ \protect \BOthers {.}}{%
{\protect \APACyear {2019}}%
}]{%
aiello2019large}
\APACinsertmetastar {%
aiello2019large}%
\begin{APACrefauthors}%
Aiello, L\BPBI M.%
, Schifanella, R.%
, Quercia, D.%
\BCBL {}\ \BBA {} Del~Prete, L.%
\end{APACrefauthors}%
\unskip\
\newblock
\APACrefYearMonthDay{2019}{}{}.
\newblock
{\BBOQ}\APACrefatitle {Large-scale and high-resolution analysis of food
  purchases and health outcomes} {Large-scale and high-resolution analysis of
  food purchases and health outcomes}.{\BBCQ}
\newblock
\APACjournalVolNumPages{EPJ Data Science}{8}{1}{14}.
\PrintBackRefs{\CurrentBib}

\bibitem [\protect \citeauthoryear {%
Amato%
\ \BBA {} Cozzolino%
}{%
Amato%
\ \BBA {} Cozzolino%
}{%
{\protect \APACyear {2020}}%
}]{%
amato2020safeeat}
\APACinsertmetastar {%
amato2020safeeat}%
\begin{APACrefauthors}%
Amato, A.%
\BCBT {}\ \BBA {} Cozzolino, G.%
\end{APACrefauthors}%
\unskip\
\newblock
\APACrefYearMonthDay{2020}{}{}.
\newblock
{\BBOQ}\APACrefatitle {SafeEat: Extraction of Information About the Presence of
  Food Allergens in Recipes} {Safeeat: Extraction of information about the
  presence of food allergens in recipes}.{\BBCQ}
\newblock
\BIn{} \APACrefbtitle {International Conference on Intelligent Networking and
  Collaborative Systems} {International conference on intelligent networking
  and collaborative systems}\ (\BPGS\ 194--203).
\PrintBackRefs{\CurrentBib}

\bibitem [\protect \citeauthoryear {%
Barto%
, Bradtke%
\BCBL {}\ \BBA {} Singh%
}{%
Barto%
\ \protect \BOthers {.}}{%
{\protect \APACyear {1995}}%
}]{%
barto1995learning}
\APACinsertmetastar {%
barto1995learning}%
\begin{APACrefauthors}%
Barto, A\BPBI G.%
, Bradtke, S\BPBI J.%
\BCBL {}\ \BBA {} Singh, S\BPBI P.%
\end{APACrefauthors}%
\unskip\
\newblock
\APACrefYearMonthDay{1995}{}{}.
\newblock
{\BBOQ}\APACrefatitle {Learning to act using real-time dynamic programming}
  {Learning to act using real-time dynamic programming}.{\BBCQ}
\newblock
\APACjournalVolNumPages{Artificial intelligence}{72}{1-2}{81--138}.
\PrintBackRefs{\CurrentBib}

\bibitem [\protect \citeauthoryear {%
Bengio%
, Courville%
\BCBL {}\ \BBA {} Vincent%
}{%
Bengio%
\ \protect \BOthers {.}}{%
{\protect \APACyear {2013}}%
}]{%
bengio2013representation}
\APACinsertmetastar {%
bengio2013representation}%
\begin{APACrefauthors}%
Bengio, Y.%
, Courville, A.%
\BCBL {}\ \BBA {} Vincent, P.%
\end{APACrefauthors}%
\unskip\
\newblock
\APACrefYearMonthDay{2013}{}{}.
\newblock
{\BBOQ}\APACrefatitle {Representation learning: A review and new perspectives}
  {Representation learning: A review and new perspectives}.{\BBCQ}
\newblock
\APACjournalVolNumPages{IEEE transactions on pattern analysis and machine
  intelligence}{35}{8}{1798--1828}.
\PrintBackRefs{\CurrentBib}

\bibitem [\protect \citeauthoryear {%
Boutsioukis%
, Partalas%
\BCBL {}\ \BBA {} Vlahavas%
}{%
Boutsioukis%
\ \protect \BOthers {.}}{%
{\protect \APACyear {2011}}%
}]{%
boutsioukis2011transfer}
\APACinsertmetastar {%
boutsioukis2011transfer}%
\begin{APACrefauthors}%
Boutsioukis, G.%
, Partalas, I.%
\BCBL {}\ \BBA {} Vlahavas, I.%
\end{APACrefauthors}%
\unskip\
\newblock
\APACrefYearMonthDay{2011}{}{}.
\newblock
{\BBOQ}\APACrefatitle {Transfer learning in multi-agent reinforcement learning
  domains} {Transfer learning in multi-agent reinforcement learning
  domains}.{\BBCQ}
\newblock
\BIn{} \APACrefbtitle {European Workshop on Reinforcement Learning} {European
  workshop on reinforcement learning}\ (\BPGS\ 249--260).
\PrintBackRefs{\CurrentBib}

\bibitem [\protect \citeauthoryear {%
Cutting%
, Kupiec%
, Pedersen%
\BCBL {}\ \BBA {} Sibun%
}{%
Cutting%
\ \protect \BOthers {.}}{%
{\protect \APACyear {1992}}%
}]{%
cutting1992practical}
\APACinsertmetastar {%
cutting1992practical}%
\begin{APACrefauthors}%
Cutting, D.%
, Kupiec, J.%
, Pedersen, J.%
\BCBL {}\ \BBA {} Sibun, P.%
\end{APACrefauthors}%
\unskip\
\newblock
\APACrefYearMonthDay{1992}{}{}.
\newblock
{\BBOQ}\APACrefatitle {A practical part-of-speech tagger} {A practical
  part-of-speech tagger}.{\BBCQ}
\newblock
\BIn{} \APACrefbtitle {Proceedings of the third conference on Applied natural
  language processing} {Proceedings of the third conference on applied natural
  language processing}\ (\BPGS\ 133--140).
\PrintBackRefs{\CurrentBib}

\bibitem [\protect \citeauthoryear {%
Fukushima%
\ \BBA {} Miyake%
}{%
Fukushima%
\ \BBA {} Miyake%
}{%
{\protect \APACyear {1982}}%
}]{%
fukushima1982neocognitron}
\APACinsertmetastar {%
fukushima1982neocognitron}%
\begin{APACrefauthors}%
Fukushima, K.%
\BCBT {}\ \BBA {} Miyake, S.%
\end{APACrefauthors}%
\unskip\
\newblock
\APACrefYearMonthDay{1982}{}{}.
\newblock
{\BBOQ}\APACrefatitle {Neocognitron: A self-organizing neural network model for
  a mechanism of visual pattern recognition} {Neocognitron: A self-organizing
  neural network model for a mechanism of visual pattern recognition}.{\BBCQ}
\newblock
\BIn{} \APACrefbtitle {Competition and cooperation in neural nets} {Competition
  and cooperation in neural nets}\ (\BPGS\ 267--285).
\newblock
\APACaddressPublisher{}{Springer}.
\PrintBackRefs{\CurrentBib}

\bibitem [\protect \citeauthoryear {%
LeCun%
, Bengio%
\BCBL {}\ \BBA {} Hinton%
}{%
LeCun%
\ \protect \BOthers {.}}{%
{\protect \APACyear {2015}}%
}]{%
lecun2015deep}
\APACinsertmetastar {%
lecun2015deep}%
\begin{APACrefauthors}%
LeCun, Y.%
, Bengio, Y.%
\BCBL {}\ \BBA {} Hinton, G.%
\end{APACrefauthors}%
\unskip\
\newblock
\APACrefYearMonthDay{2015}{}{}.
\newblock
{\BBOQ}\APACrefatitle {Deep learning} {Deep learning}.{\BBCQ}
\newblock
\APACjournalVolNumPages{nature}{521}{7553}{436--444}.
\PrintBackRefs{\CurrentBib}

\bibitem [\protect \citeauthoryear {%
LeCun%
\ \protect \BOthers {.}}{%
LeCun%
\ \protect \BOthers {.}}{%
{\protect \APACyear {1989}}%
}]{%
lecun1989backpropagation}
\APACinsertmetastar {%
lecun1989backpropagation}%
\begin{APACrefauthors}%
LeCun, Y.%
, Boser, B.%
, Denker, J\BPBI S.%
, Henderson, D.%
, Howard, R\BPBI E.%
, Hubbard, W.%
\BCBL {}\ \BBA {} Jackel, L\BPBI D.%
\end{APACrefauthors}%
\unskip\
\newblock
\APACrefYearMonthDay{1989}{}{}.
\newblock
{\BBOQ}\APACrefatitle {Backpropagation applied to handwritten zip code
  recognition} {Backpropagation applied to handwritten zip code
  recognition}.{\BBCQ}
\newblock
\APACjournalVolNumPages{Neural computation}{1}{4}{541--551}.
\PrintBackRefs{\CurrentBib}

\bibitem [\protect \citeauthoryear {%
Mikheev%
}{%
Mikheev%
}{%
{\protect \APACyear {1996}}%
}]{%
mikheev1996learning}
\APACinsertmetastar {%
mikheev1996learning}%
\begin{APACrefauthors}%
Mikheev, A.%
\end{APACrefauthors}%
\unskip\
\newblock
\APACrefYearMonthDay{1996}{}{}.
\newblock
{\BBOQ}\APACrefatitle {Learning part-of-speech guessing rules from lexicon:
  Extension to non-concatenative operations} {Learning part-of-speech guessing
  rules from lexicon: Extension to non-concatenative operations}.{\BBCQ}
\newblock
\BIn{} \APACrefbtitle {Proceedings of the 16th conference on Computational
  linguistics-Volume 2} {Proceedings of the 16th conference on computational
  linguistics-volume 2}\ (\BPGS\ 770--775).
\PrintBackRefs{\CurrentBib}

\bibitem [\protect \citeauthoryear {%
Mnih%
\ \protect \BOthers {.}}{%
Mnih%
\ \protect \BOthers {.}}{%
{\protect \APACyear {2015}}%
}]{%
mnih2015human}
\APACinsertmetastar {%
mnih2015human}%
\begin{APACrefauthors}%
Mnih, V.%
, Kavukcuoglu, K.%
, Silver, D.%
, Rusu, A\BPBI A.%
, Veness, J.%
, Bellemare, M\BPBI G.%
\BDBL {}others%
\end{APACrefauthors}%
\unskip\
\newblock
\APACrefYearMonthDay{2015}{}{}.
\newblock
{\BBOQ}\APACrefatitle {Human-level control through deep reinforcement learning}
  {Human-level control through deep reinforcement learning}.{\BBCQ}
\newblock
\APACjournalVolNumPages{nature}{518}{7540}{529--533}.
\PrintBackRefs{\CurrentBib}

\bibitem [\protect \citeauthoryear {%
Rabiner%
}{%
Rabiner%
}{%
{\protect \APACyear {1989}}%
}]{%
rabiner1989tutorial}
\APACinsertmetastar {%
rabiner1989tutorial}%
\begin{APACrefauthors}%
Rabiner, L\BPBI R.%
\end{APACrefauthors}%
\unskip\
\newblock
\APACrefYearMonthDay{1989}{}{}.
\newblock
{\BBOQ}\APACrefatitle {A tutorial on hidden Markov models and selected
  applications in speech recognition} {A tutorial on hidden markov models and
  selected applications in speech recognition}.{\BBCQ}
\newblock
\APACjournalVolNumPages{Proceedings of the IEEE}{77}{2}{257--286}.
\PrintBackRefs{\CurrentBib}

\bibitem [\protect \citeauthoryear {%
Sajadmanesh%
\ \protect \BOthers {.}}{%
Sajadmanesh%
\ \protect \BOthers {.}}{%
{\protect \APACyear {2017}}%
}]{%
sajadmanesh2017kissing}
\APACinsertmetastar {%
sajadmanesh2017kissing}%
\begin{APACrefauthors}%
Sajadmanesh, S.%
, Jafarzadeh, S.%
, Ossia, S\BPBI A.%
, Rabiee, H\BPBI R.%
, Haddadi, H.%
, Mejova, Y.%
\BDBL {}Stringhini, G.%
\end{APACrefauthors}%
\unskip\
\newblock
\APACrefYearMonthDay{2017}{}{}.
\newblock
{\BBOQ}\APACrefatitle {Kissing cuisines: Exploring worldwide culinary habits on
  the web} {Kissing cuisines: Exploring worldwide culinary habits on the
  web}.{\BBCQ}
\newblock
\BIn{} \APACrefbtitle {Proceedings of the 26th International Conference on
  World Wide Web Companion} {Proceedings of the 26th international conference
  on world wide web companion}\ (\BPGS\ 1013--1021).
\PrintBackRefs{\CurrentBib}

\bibitem [\protect \citeauthoryear {%
Silva%
, Ribeiro%
\BCBL {}\ \BBA {} Ferreira%
}{%
Silva%
\ \protect \BOthers {.}}{%
{\protect \APACyear {2019}}%
}]{%
silva2019information}
\APACinsertmetastar {%
silva2019information}%
\begin{APACrefauthors}%
Silva, N.%
, Ribeiro, D.%
\BCBL {}\ \BBA {} Ferreira, L.%
\end{APACrefauthors}%
\unskip\
\newblock
\APACrefYearMonthDay{2019}{}{}.
\newblock
{\BBOQ}\APACrefatitle {Information extraction from unstructured recipe data}
  {Information extraction from unstructured recipe data}.{\BBCQ}
\newblock
\BIn{} \APACrefbtitle {Proceedings of the 2019 5th International Conference on
  Computer and Technology Applications} {Proceedings of the 2019 5th
  international conference on computer and technology applications}\ (\BPGS\
  165--168).
\PrintBackRefs{\CurrentBib}

\bibitem [\protect \citeauthoryear {%
Simas%
, Ficek%
, Diaz-Guilera%
, Obrador%
\BCBL {}\ \BBA {} Rodriguez%
}{%
Simas%
\ \protect \BOthers {.}}{%
{\protect \APACyear {2017}}%
}]{%
simas2017food}
\APACinsertmetastar {%
simas2017food}%
\begin{APACrefauthors}%
Simas, T.%
, Ficek, M.%
, Diaz-Guilera, A.%
, Obrador, P.%
\BCBL {}\ \BBA {} Rodriguez, P\BPBI R.%
\end{APACrefauthors}%
\unskip\
\newblock
\APACrefYearMonthDay{2017}{}{}.
\newblock
{\BBOQ}\APACrefatitle {Food-bridging: a new network construction to unveil the
  principles of cooking} {Food-bridging: a new network construction to unveil
  the principles of cooking}.{\BBCQ}
\newblock
\APACjournalVolNumPages{Frontiers in ICT}{4}{}{14}.
\PrintBackRefs{\CurrentBib}

\bibitem [\protect \citeauthoryear {%
Thede%
\ \BBA {} Harper%
}{%
Thede%
\ \BBA {} Harper%
}{%
{\protect \APACyear {1999}}%
}]{%
thede1999second}
\APACinsertmetastar {%
thede1999second}%
\begin{APACrefauthors}%
Thede, S\BPBI M.%
\BCBT {}\ \BBA {} Harper, M\BPBI P.%
\end{APACrefauthors}%
\unskip\
\newblock
\APACrefYearMonthDay{1999}{}{}.
\newblock
{\BBOQ}\APACrefatitle {A second-order hidden Markov model for part-of-speech
  tagging} {A second-order hidden markov model for part-of-speech
  tagging}.{\BBCQ}
\newblock
\BIn{} \APACrefbtitle {Proceedings of the 37th annual meeting of the
  Association for Computational Linguistics on Computational Linguistics}
  {Proceedings of the 37th annual meeting of the association for computational
  linguistics on computational linguistics}\ (\BPGS\ 175--182).
\PrintBackRefs{\CurrentBib}

\bibitem [\protect \citeauthoryear {%
Van~Erp%
\ \protect \BOthers {.}}{%
Van~Erp%
\ \protect \BOthers {.}}{%
{\protect \APACyear {2021}}%
}]{%
van2021using}
\APACinsertmetastar {%
van2021using}%
\begin{APACrefauthors}%
Van~Erp, M.%
, Reynolds, C.%
, Maynard, D.%
, Starke, A.%
, Ib{\'a}{\~n}ez~Mart{\'\i}n, R.%
, Andres, F.%
\BDBL {}others%
\end{APACrefauthors}%
\unskip\
\newblock
\APACrefYearMonthDay{2021}{}{}.
\newblock
{\BBOQ}\APACrefatitle {Using Natural Language Processing and Artificial
  Intelligence to explore the nutrition and sustainability of recipes and food}
  {Using natural language processing and artificial intelligence to explore the
  nutrition and sustainability of recipes and food}.{\BBCQ}
\newblock
\APACjournalVolNumPages{Frontiers in Artificial Intelligence}{3}{621577}{}.
\PrintBackRefs{\CurrentBib}

\bibitem [\protect \citeauthoryear {%
Vithayathil~Varghese%
\ \BBA {} Mahmoud%
}{%
Vithayathil~Varghese%
\ \BBA {} Mahmoud%
}{%
{\protect \APACyear {2020}}%
}]{%
vithayathil2020survey}
\APACinsertmetastar {%
vithayathil2020survey}%
\begin{APACrefauthors}%
Vithayathil~Varghese, N.%
\BCBT {}\ \BBA {} Mahmoud, Q\BPBI H.%
\end{APACrefauthors}%
\unskip\
\newblock
\APACrefYearMonthDay{2020}{}{}.
\newblock
{\BBOQ}\APACrefatitle {A survey of multi-task deep reinforcement learning} {A
  survey of multi-task deep reinforcement learning}.{\BBCQ}
\newblock
\APACjournalVolNumPages{Electronics}{9}{9}{1363}.
\PrintBackRefs{\CurrentBib}

\bibitem [\protect \citeauthoryear {%
Zhou%
\ \BBA {} Su%
}{%
Zhou%
\ \BBA {} Su%
}{%
{\protect \APACyear {2002}}%
}]{%
zhou2002named}
\APACinsertmetastar {%
zhou2002named}%
\begin{APACrefauthors}%
Zhou, G.%
\BCBT {}\ \BBA {} Su, J.%
\end{APACrefauthors}%
\unskip\
\newblock
\APACrefYearMonthDay{2002}{}{}.
\newblock
{\BBOQ}\APACrefatitle {Named entity recognition using an HMM-based chunk
  tagger} {Named entity recognition using an hmm-based chunk tagger}.{\BBCQ}
\newblock
\BIn{} \APACrefbtitle {proceedings of the 40th Annual Meeting on Association
  for Computational Linguistics} {proceedings of the 40th annual meeting on
  association for computational linguistics}\ (\BPGS\ 473--480).
\PrintBackRefs{\CurrentBib}

\end{thebibliography}
\end{document}